\documentclass[conference]{IEEEtran}
\IEEEoverridecommandlockouts
\usepackage{cite}
\usepackage{amsmath,amssymb,amsfonts}
\usepackage{algorithmic}
\usepackage{graphicx}
\usepackage{textcomp}
\usepackage{xcolor}
\def\BibTeX{{\rm B\kern-.05em{\sc i\kern-.025em b}\kern-.08em
    T\kern-.1667em\lower.7ex\hbox{E}\kern-.125emX}}
\begin{document}

\title{MA-NeRF: Motion-Assisted Neural Radiance Fields for Face Synthesis from Sparse Images\\
}

\author{\IEEEauthorblockN{
Weichen Zhang$^1$\IEEEauthorrefmark{1}\thanks{\IEEEauthorrefmark{1}Equal contribution}, 
Xiang Zhou$^1$\IEEEauthorrefmark{1}, 
Yukang Cao$^2$, 
Wensen Feng$^3$\IEEEauthorrefmark{2}
and Chun Yuan $^1$\IEEEauthorrefmark{2}\thanks{\IEEEauthorrefmark{2}Corresponding author}}
\IEEEauthorblockA{
$^1$Shenzhen International Graduate School, Tsinghua University\\
Shenzhen 518055, China\\
Email: \{zwc21,zhoux21\}@mails.tsinghua.edu.cn;yuanc@sz.tsinghua.edu.cn}
\IEEEauthorblockA{$^2$The University of Hong Kong\\
HongKong, China \\
Email: ykcao@cs.hku.hk}
\IEEEauthorblockA{$^3$CG \& XR Department, Huawei Technologies\\
Shenzhen 518129, China\\
Email: fengwensen@huawei.com}
}

\maketitle

\begin{abstract}
We address the problem of photorealistic 3D face avatar synthesis from sparse images. 
Existing Parametric models for face avatar reconstruction struggle to generate details that originate from inputs. 
Meanwhile, although current NeRF-based avatar methods provide promising results for novel view synthesis, they fail to generalize well for unseen expressions.
We improve from NeRF and propose a novel framework that, by leveraging the parametric 3DMM models, can reconstruct a high-fidelity drivable face avatar and successfully handle the unseen expressions. At the core of our implementation are structured displacement feature and semantic-aware learning module. Our structured displacement feature will introduce the motion prior as an additional constraints and help perform better for unseen expressions, by constructing displacement volume. Besides, the semantic-aware learning incorporates multi-level prior, e.g., semantic embedding, learnable latent code, to lift the performance to a higher level.
Thorough experiments have been doen both quantitatively and qualitatively to demonstrate the design of our framework, and our method achieves much better results than the current state-of-the-arts.
\end{abstract}

\section{Introduction}
\label{sec:intro}
Synthesizing photorealistic 3D face avatar from monocular video is an attractive topic with considerable atttention. We can find its increeasing demand in the industry applications, e.g., VR/AR utility, metaverse\cite{cao2022authentic}, etc. Early works provide promising results under motion capturing system, such as AVATAR by James Cameron, but requiring expensive equipments and complicated polishments. With the huge success of deep learning techonology, 3D Morphable Model (3DMM)~\cite{blanz1999morphable} is introduced to reconstruct a 3D face model by regressing the expression, geometry, and texture parameters. However, 3DMM-based methods exclude the hair topology and face accessories like glasses. Unacceptable rendering result is another fatal drawback which limits its application for face avatar synthesis. 

Recently, implicit-function approaches arise to provide dominating results, introducing expressive reconstruction of 3D face avatar with realistic rendering details, and sharper geometry topologies\cite{zheng2022avatar,GuyGafni2021DynamicNR,AlbertPumarola2020DNeRFNR}. As one of the most representative methods, Neural Radiance Field (NeRF)\cite{mildenhall2021nerf} optimizes a Multilayer Perceptron (MLP) on the given monocular video to predict the density and color and accomplishes novel view synthesis. However, despite the promising results introduced by NeRF and follow-up methods\cite{AlbertPumarola2020DNeRFNR,park2021hypernerf,park2021nerfies}, they encounter two major problems that will drastically pull down the performance: 
(1) It’s an extremely under-constrained problem to optimize translational vector fields in NeRF.
(2) They struggle with unseen expressions and will reconstruct notable artifacts. These two phenomena will become worse with smaller training dataset. 

In this paper, to address the ill-posed optimization problem and to make our network well-generalized for unseen expressions, we hereby propose a novel motion-assisted neural radiance fields for face synthesis model(so-called MA-NeRF) from sparse images. Specifically, MA-NeRF predicts the desity and color value to accomplish image rendering from a sparse set of images. To explicitly control the face avatar, we utilize the multi-level prior from 3DMM, i.e., expression, semantic embedding, displacement and latent codes, so that we are enabled to synthesize a better expression under sparse inputs. we further propose our novel structured displacement feature and semantic-aware learning module, which allow us to achieve generalizable face avatar synthesis. 

We evaluate the proposed MA-NeRF on multiple public datasets both qualitatively and quantitatively. Our method achieves impressive performance under both full input setting and sparse input setting, surpassing previous state-of-the-art methods. To summarize, the contributions of the paper are as follows:
\begin{itemize}
    \item We propose MA-NeRF that, by incorporating 3DMM model as 3D prior, is capable of reconstructing a high-fidelity drivable avatar under sparse image inputs.
    \item We incorporate structured displacement feature and semantic-aware learning module to achieve phenomenal results on unseen expressions.
    \item We conduct extensive experiments on public datasets and achieve significant improvements over previous state-of-the-art methods.
\end{itemize}
\section{Related Work}

\subsection{3D Morphable Models}
3D morphable face model(3DMM)\cite{blanz1999morphable} is a generic 3D face model representing a face with a fixed number of vertices. Its core idea is to learn low-dimensional priors of facial geometry, appearance, and expressions from a large corpus of high-fidelity face reconstructions. Unfortunately, although 3DMM and its variants have been widely used within optimization-based\cite{gecer2019ganfit} and deep learning-based approaches\cite{feng2021learning}, the features derived from the low-dimensional priors prevent the modelling of person-specific traits such as hair and wrinkle-level details. \cite{cao2022authentic}. In our work, we model each face individually using a novel implicit representation, resulting in improved rendering outcomes for facial images.
\subsection{Neural Face Models}
NeRF and its extensions\cite{mildenhall2021nerf,martin2021nerf} enable high quality rendering of novel views of static scenes. Compared with 3DMM, they can model more personalized and complex details, such as hair and glasses. Some work extended NeRF to dynamic human face modelling.\cite{GuyGafni2021DynamicNR} However, this work demonstrates strong performance in terms of image quality for novel views and interpolated expressions; it falls short in terms of extrapolation to unseen expressions because of the lack of a motion prior. In Our work, we use the rough 3D facial information from 3DMM, calculating the displacement of each vertex from canonical space to a specific expression as the motion prior of the neural radiation field. It also constrains the neural radiation field to learn a shared canonical geometry for different expressions. 

\section{Method}
\begin{figure*}[tbp]
	\centering
	\includegraphics[width=1\textwidth]{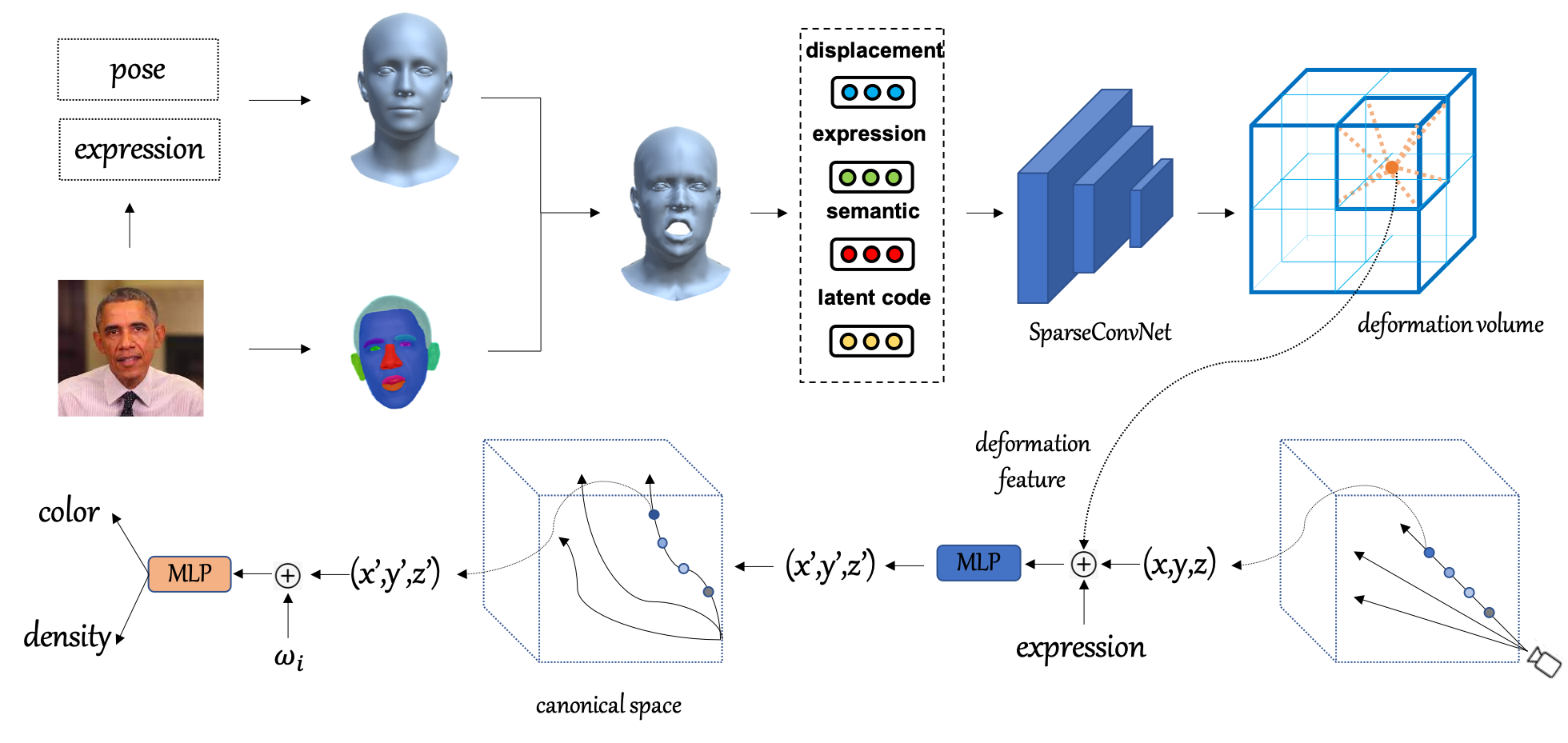}
	\vspace{-1cm}
	\caption{The architecture of MA-NeRF. For each vertex of the 3DMM, we obtain the displacement by explicit deformation using expression parameters, and fusion displacement with semantic features from semantic image, latent code and shared expression parameters, and input them into SparseConvNet together to obtain the deformation volume. For any point x in 3D space, we query its deformation feature from the deformation volume and maps 3D points to a canonical space for rendering.
	}
        \label{fig:pipeline}
\end{figure*}
We introduce \textbf{MA-NeRF}: a motion-assisted neural radiance field for face synthesis from sparse images. The core techniques of our framework are a structured displacement feature and a semantic-aware learning module. Figure~\ref{fig:pipeline} provides an overview of our framework. In the following sections, we will first brief on the basic knowledge of the proposed MA-NeRF in Sec.\ref{sec:preliminary}. After that, we will introduce our basic dynamic neural radiance module in Sec.\ref{sec:nerf}. We will then describe how we leverage the motion prior from 3DMM to guide the deformation of observed points from the observation space to the canonical space in Sec.\ref{sec:deformation}, followed by the description of how we learn structured semantic information through semantic graph projection in Sec.\ref{sec:semantic}.  Finally, we will discuss  some training details in Sec.\ref{sec:training}.

\subsection{Preliminary}
\label{sec:preliminary}
\noindent
\textbf{Neural Radiance Field (NeRF). }

Neural radiance field (NeRF) is proposed to generate an rendered image via a continuous and volumetric representation. Overall it maps a point $\mathbf{x} = (x, y, z)$ in 3D space to the corresponding color $\mathbf{c} = (r, g, b)$ and volume density value $\sigma$:  
\begin{eqnarray}
    F(\mathbf{x},\mathbf{d})\rightarrow(\mathbf{c},\sigma),
\end{eqnarray}
where $\mathbf{d}$ is the ray direction.

Volume rendering is the main technique proposed in NeRF to render the implicit geometry and appearance representation onto a 2D image. During the rendering procedure, we will integrate the sampled density and RGB values along the rays $\mathbf{d}$ cast through each pixel. The rendered color $\hat{C}$ of the each pixel in the 2D image space is given by:

\begin{eqnarray}
\begin{array}{c}\hat{C}(\mathbf{r})=\sum_{k=1}^{N_{k}} T_{k}\left(1-\exp \left(-\sigma_{t}\left(\mathbf{x}_{k}\right) \delta_{k}\right)\right) \mathbf{c}\left(\mathbf{x}_{k}\right), \\ \text { where } \quad T_{k}=\exp \left(-\sum_{j=1}^{k-1} \sigma\left(\mathbf{x}_{j}\right) \delta_{j}\right)\end{array},
\end{eqnarray}
where $\delta_{k}=\left\|\mathbf{x}_{k+1}-\mathbf{x}_{k}\right\|_{2}$ is the distance between adjacent sampled points. We set $N_k$ as 64 in our experiments. 

\noindent
\textbf{3D Morphable Model (3DMM). }
We specifically choose 3DMM as our 3D prior considering its robust results. 3DMM\cite{blanz1999morphable} reconstructs 3D face model by regressing the expression parameters $\psi$, identity parameters $\beta$, shape, pose and texture parameters from an extensive training dataset, which is collected from widely ranged populations. 
\begin{eqnarray}
\begin{aligned} 
V_{\text{3DMM}}: \mathbb{R}^{99}, \mathbb{R}^{79} &\rightarrow \mathbb{R}^{34650 \times 3}\\
\beta, \psi & \mapsto V_{\text {3DMM}}(\beta, \psi)
\end{aligned}
\end{eqnarray}
Displacement is one of the essential components of 3DMM networks, and will serve as the basis for our structured displacement feature. Generally, we can obtain the displacement information by calculating per vertex difference with the standard 3DMM model.
\begin{eqnarray}
\Delta V_{\text {3DMM}} &= V_{\text {3DMM}}(\beta, \psi_{can}) - V_{\text {3DMM}}(\beta, \psi)
\end{eqnarray}
\subsection{Dynamic Neural Radiance Fields}
\label{sec:nerf}
To acquire an efficient dynamic model, we decoupled the rigid and non-rigid aspects of deformation. The overall head pose is considered to be a rigid-body transformation, while specific expression changes are treated as non-rigid deformation.

Similar to \cite{GuyGafni2021DynamicNR}, we utilize changes of camera angle to simulate the rigid-body transformation of the human head. The pose parameters, including rotation and translation, from the face tracking algorithm are used to deform the rays into a canonical space. Note that the canonical space remains unchanged across all frames.

For non-rigid deformation, e.g., face expressions, we apply a dynamic face model by combining the canonical face neural radiance field and a deformation field, inspired by \cite{park2021nerfies,AlbertPumarola2020DNeRFNR,park2021hypernerf}. Specifically, given an observation-space point x, the deformation field transforms x to the canonical space, conditioned on the expression parameters $\psi_i$.

Now, our \textbf{basic} implicit function of deformation field can be formulated as: 

\begin{eqnarray}
F^{\text {deform }}:\left(\mathbf{x},\psi_i\right) \rightarrow \Delta \mathbf{x},
\end{eqnarray}
where $i$ represents the $i$th frame.

\noindent
\textbf{Canonical facial model. }

We use MLP to regress the color and density of the corresponding point $x+\Delta x$ of $x$ in the canonical space.
We apply the positional encoding to both the viewing direction $d$ and the spatial location $x$, which enables better learning of high frequency functions. 

Following NeRF-W\cite{martin2021nerf}, we incorporate a per-frame appearance latent code $\omega_i$, which equips us to adjust the color output. Therefore, we are able to address the variations among different frames, e.g., the exposure difference and variations of the white balance, and to compensate for the errors in face expression and pose estimation at the same time. 
The model is defined as:
\begin{eqnarray}
F_{\Theta}:(\mathbf{x}+\Delta \mathbf{x}, \mathbf{d},\omega_i) \rightarrow(\mathbf{c}, \sigma)
\end{eqnarray}
\subsection{Structured Displacement Feature}
\label{sec:deformation}

We found that deforming only conditioned on the expression parameters made it difficult for the model to model unseen expression parameters, especially when the input image is relatively sparse. Therefore, we use the vertex displacement variation of the 3DMM as a motion prior to assiste the deformation of the neural radiation field.

One naive way to obtain the displacement is directly using the 3DMM inner calculation. However, this displacement of vertices brought by 3DMM has two major drawbacks: (1) it is discrete, while we need a continuous space in the neural radiance field. (2) It only includes the skin of the face, and lacks the head topologies, e.g., hair, glasses.

Inspired by \cite{SongyouPeng2020ConvolutionalON, peng2021neural}, we formulate a displacement volume by anchoring and diffusing the displacement of 3DMM with SparseConvNet~\cite{BenjaminGraham20173DSS}. With this displacement volume, for a point $\mathbf{x}$ in the 3D space, we can continuously query its displacement feature with tri-linear interpolation. Our 3DMM-guided structured displacement feature of a 3D point $\mathbf{x}$ can be then formulated as:

\begin{eqnarray}
f = \mathcal{T}(x,\mathcal{D}_{\phi}(\psi,\Delta V_{\text {3DMM}})),
\end{eqnarray}
where $\Delta V_{\text {3DMM}} = \{\Delta v_{1},\Delta v_{2},\cdots,\Delta v_{34650}\}$ is the displacement by 3DMM, $\psi$ is the 3DMM expression parameters, $\phi$ represents the parameters of the SparseConvNet $\mathcal{D}_{\phi}$, and $\mathcal{T}$ means trilinear interpolation.

\subsection{Semantic-Aware Learning}
\label{sec:semantic}
Considering that each sementic information, e.g., mouth, eyes, nose, of the avatar indidates different appearances and moving patterns, we should therefore handle them differently even with the same expression~\cite{liu2022semantic}.
However, the expression parameters directly obtained from 3DMM cannot distinguish different face semantics. To handle this problem, (1) for each vertex of the canonical 3DMM mesh, we project it into the canonical semantic graph.
(2) We anchor a set of learnable latent codes to the vertices of the canonical 3DMM to learn more detailed structural information.

Together with the structured displacement feature, we obtain the final structured deformation feature via SparseConvNet: 
\begin{eqnarray}
f^{\prime} = \mathcal{T}(x,\mathcal{D}_{\phi}(\psi,\Delta V_{\text {3DMM}},\mathcal{S},\mathcal{Z}))
\end{eqnarray}
where $\mathcal{S} = \{s_{1},s_{2},\cdots,s_{34650}\}$ is the semantic category obtained by projecting each vertex onto the canonical semantic graph and $\mathcal{Z} = \{z_{1},z_{2},\cdots,z_{34650}\}$ is a set of latent code on vertices of the 3DMM.

Till Now, we can update our implicit function of deformation field in Sec.\ref{sec:nerf} with final structured deformation feature $f^{\prime}_{i}$ as: 
\begin{eqnarray}
F^{\text {deform }}:\left(\mathbf{x},\psi_i,f^{\prime}_i\right) \rightarrow \Delta \mathbf{x},
\end{eqnarray}
where $i$ represents the $i$th frame.

\subsection{Network Training}
\label{sec:training}
\label{loss}
Similar to NeRF which simultaneously optimizes coarse and fine models with hierarchical volume rendering, we train the network with the following photometric loss $\mathcal{L}_{p}$:
\begin{eqnarray}
\mathcal{L}_{p}=\sum_{\mathbf{r} \in \mathcal{R}}\left[\left\|\hat{C}_{c}\left(\mathbf{r}\right)-C\left(\mathbf{r}\right)\right\|_{2}^{2}+\left\|\hat{C}_{f}\left(\mathbf{r}\right)-C\left(\mathbf{r}\right)\right\|_{2}^{2}\right]
\end{eqnarray}
where $\mathcal{R}$ is the set of camera rays passing through image pixels, and $C(r)$ means the ground-truth pixel color. 

In addition, we employ a hard surface $\mathcal{L}_{hard}$ to reduce the halo effect surrounding the canonical face. Specifically, we encourage that the weight of each sample be either 1 or 0 given by,
\begin{eqnarray}
\mathcal{L}_{hard}=-\log \left(e^{-|w|}+e^{-|1-w|}\right)
\end{eqnarray}
where $w$ refers to the transparency where the ray terminates .

However, this penalty alone is not enough to obtain a sharp canonical shape, we also add a canonical edge loss, $\mathcal{L}_{edge}$. By rendering a random straight ray in the canonical volume, we encourage the accumulated alpha values to be either 1 or 0. This is given by,
\begin{eqnarray}
\mathcal{L}_{edge}=-\log \left(e^{-|\alpha_c|}+e^{-|1-\alpha_c|}\right)
\end{eqnarray}
where $\alpha_c$ is the accumulated alpha value obtained from a random straight ray in canonical space.

The overall learning objective for the framework is:
\begin{eqnarray}
\mathcal{L}=\mathcal{L}_p + \lambda_1 \mathcal{L}_{hard} +\lambda_2 \mathcal{L}_{edge} 
\end{eqnarray}
where $\lambda_1$ and $\lambda_2$ is the weight balancing coefficient.
\section{Experiment}
We hereby evaluate the results of our proposed MA-NeRF method. We provide both qualitatively and quantitatively comparison between our method with two state-of-the-art methods.
\subsection{Dataset and Preprocessing}
We evaluate on real video from a single stationary camera. 
In particular, we use publicly-released video set of AD-NeRF\cite{guo2021ad} and NerFACE\cite{GuyGafni2021DynamicNR}, which from collected videos on the Internet or self-collected videos.
Given the dataset, we will finish several preprocessing offline: (1) we first apply MODNet~\cite{ke2022modnet} to acquire the foreground mask. (2) We then offline estimate the head pose parameters and expression parameters by ~\cite{guo2021ad}
The estimated head pose will be used to simulate the rigid-body transformation. (3) We obtain the parsing maps of total 18 semantic classes via BiSeNet \cite{yu2018bisenet}. Different from NerFACE\cite{GuyGafni2021DynamicNR}, we take a sparse set of images (about 30-80 frames) as input and use 25 images for testing for each subject.

\noindent
\textbf{Implementation Details. }
We implement our framework with PyTorch. The networks are trained via Adam optimizer with a learning rate of 0.0005. In our experiments, we use 225 $\times$ 225 images and train each model for 200k iterations.
\subsection{Results on Real Dataset}

\begin{table}
\caption{Quantitaive evaluation for our method in comparison to other facial reconstruction methods on a sparse set of input images. Our results outperformed Neural Body\cite{peng2021neural} (based on self-reenactment for facial reconstruction), NerFACE\cite{GuyGafni2021DynamicNR} on most metrics.} 
\label{tab:results}
\begin{centering}
{
\begin{tabular}{lcccc}
\hline & PSNR \(\uparrow\) & SSIM \(\uparrow\) & LPIPS \(\downarrow\) \ & vggLPIPS \(\downarrow\) \\ 
\hline  
Neural Body & \(22.54\) & \(0.890\) & \(0.096\) & \(0.196\) \\
NerFACE & \(23.21\) & \(0.912\) & \(0.059\) & \(0.139\) \\
\hline
Ours & \(\mathbf{24.08}\) & \(\mathbf{0 . 923}\) & \(\mathbf{0 . 042}\) & \(\mathbf{0 . 128}\) \\ 
\hline
\end{tabular}
}
\end{centering}
\end{table}
\begin{figure}[ht]
	\centering
	\includegraphics[width=0.45\textwidth]{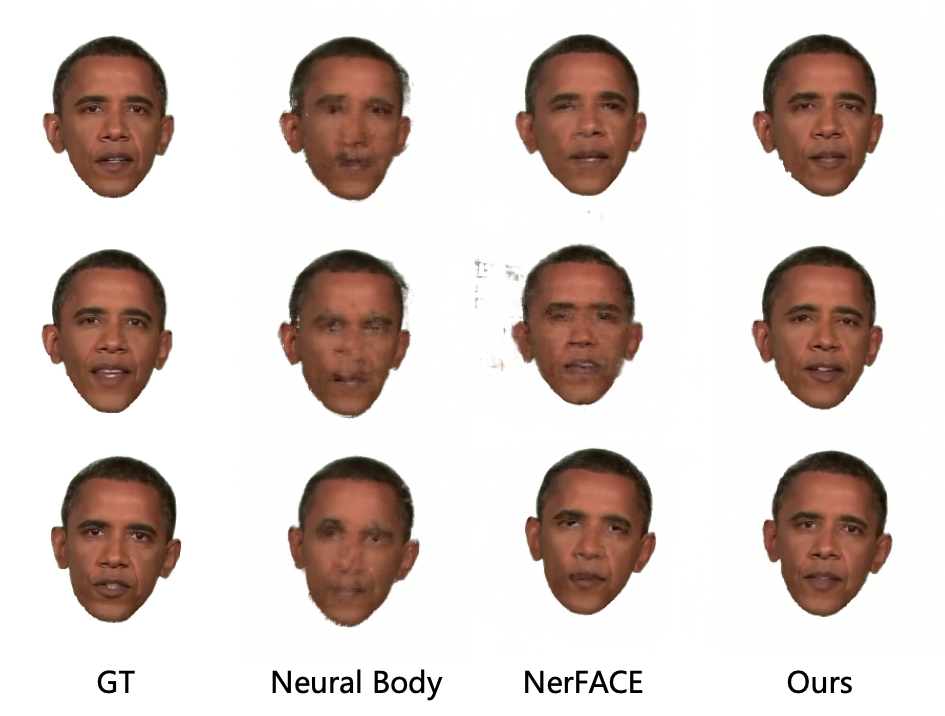}
	\caption{Qualitative comparision on a sparse set of input images. Benefiting from the motion prior and structured semantic features provided by 3DMM, MA-NeRF can generate images of higher quality compared to other methods.
	}
        \label{fig:results}
\end{figure}
We qualitatively and quantitatively compare our method with two other methods that perform closely related tasks:
(1) Neural Body\cite{peng2021neural}, using a set of latent codes anchored on the SMPL vertices to build local implicit representations. For fair comparison, we reimplement Neural Body and use 3DMM rather than SMPL to reconstruct face avatar,
(2) NerFACE\cite{GuyGafni2021DynamicNR}, a state-of-the-art open source method using NeRF for face dynamics control with expression parameters. We train official NerFACE released code on our training dataset for fair comparison.

\noindent
\textbf{Quantitative Comparison}
To quantitatively evaluate our method and the other two approaches, we compute the Peak Signal-to-Noise Ratio (PSNR), and Structure Similarity Index (SSIM)\cite{wang2004image}, as well as the Learned Perceptual Image Patch Similarity (LPIPS)\cite{zhang2018unreasonable}. Tab. \ref{tab:results} provides the detailed quantitative numbers. As can be seen our method achieves the best results for all the evaluated metrics, and even our method surpasses the others by a large margin.

\noindent
\textbf{Qualitative Comparison}
Fig. \ref{fig:results} provides the qualitative results of each method on a sparse set of input images. As can be observed, our method provides the best image quality.
NerFACE 
does not enforce a shared canonical geometry for different expressions. and due to the lack of motion prior, it is can only fit well to the expression that has been seen in the training dataset. These lead to collapse when NerFACE encounters unseen expressions. 
Neural Body largely relies on the quality of the coarse geometry (3DMM) because it directly utilizes the 3DMM vertex to establish the local implicit field. Like NerFACE, Neural Body does not establish a shared canonical geometry. These factors all lead to poor generalization of the model under unseen expressions and significant artifacts. 

\begin{figure}[h]
	\centering
	\includegraphics[width=0.5\textwidth]{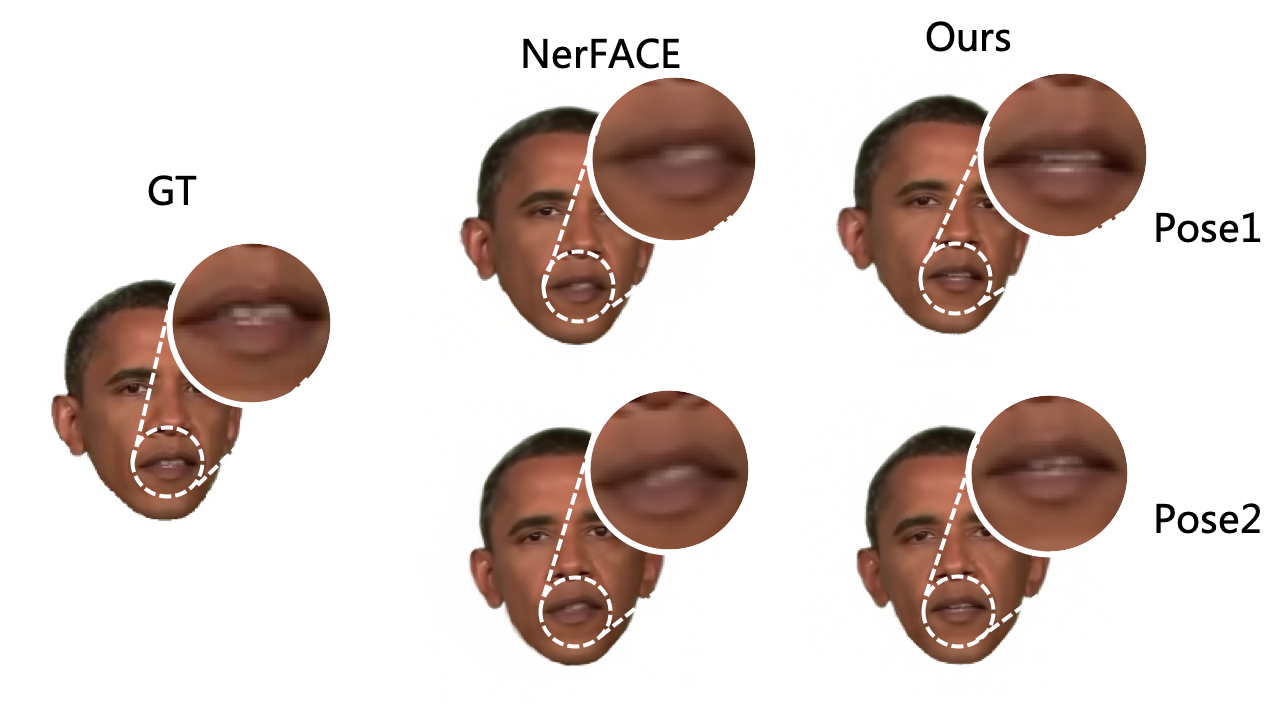}
	\caption{Qualitative comparision with more input images. The image on the far left is the expression we need to model. The image also shows the model's ability to capture the same expression under a new pose.
	}
        \label{fig:full}
\end{figure}

We further demonstrate the comparison of our model with NerFACE under the input of more images in Fig. \ref{fig:full}. Our model can establish more accurate results on unseen expressions and reduce the artifacts.

\subsection{Ablation Study}
\label{sec:ablation}
We conduct ablation studies to investigate and demonstrate the usefulness of the key aspects employed by our method (see Tab. \ref{tab:ablation}). For "w/o Semantic", we remove the semantic feature for each vertex in 3DMM which proposed in Sec. \ref{sec:semantic}. For "w/o Expression", we remove the expression parameters anchored at vertices of 3DMM. For "w/o $\mathcal{L}_{hard} $\& $\mathcal{L}_{edge}$", we only use photometric loss to supervise the training of the entire network, without the hard surface loss and canonical edge loss mentioned in Sec.\ref{loss}. 

Tab. \ref{tab:ablation} demonstrates that the semantic awareness of the model, which has been developed through learning structured semantic and expression features, enables it to better understand the appearance and geometry of each component. Simultaneously, applying increased canonical edge and hard surface loss effectively removes the undesirable fog within the volume and reduces the halo appearance in the final renderings.
\begin{table}
\caption{Ablation study. See Sec.\ref{sec:ablation} for detailed descriptions. } 
\label{tab:ablation}
\begin{centering}
{
\begin{tabular}{lcccc}
\hline & PSNR \(\uparrow\) & SSIM \(\uparrow\) & LPIPS \(\downarrow\) \ & vggLPIPS \(\downarrow\) \\ 
\hline  
w/o Semantic & \(23.60\) & \(0.915\) & \(0.048\) & \(0.144\) \\
w/o Expression & \(23.87\) & \(0.920\) & \(0.044\) & \(0.139\) \\
w/o $\mathcal{L}_{hard} $\& $\mathcal{L}_{edge}$  & \(23.89\) & \(0.917\) & \(0.044\) & \(0.135\) \\
\hline
Ours  & \(\mathbf{24.08}\) & \(\mathbf{0 . 923}\) & \(\mathbf{0 . 042}\) & \(\mathbf{0 . 128}\) \\ 
\hline
\end{tabular}
}
\end{centering}
\end{table}
\section{Conclusion}
We propose MA-NeRF, a motion-assisted facial nerf model which can learn and render controllable 3D facial avatars from sparse images. MA-NeRF uses the displacement of 3DMM vertices as motion prior to guild neural radiance fields deformation, allowing us to effectively model non-rigid changes caused by expression parameters. However, our work also has some limitations, It takes us several seconds to render a frame, and we cannot perform further editable operations on the modelled face.
\section*{Acknowledgements}
\hyphenpenalty=5000
\tolerance=1000
This work was supported by the National Key R\&D Program of China (2022YFB4701400/4701402), SZSTC Grant(JCYJ20190809172201639, WDZC20200820200655001),
Shenzhen Key Laboratory (ZDSYS20210623092001004).

\bibliographystyle{IEEEbib}
\bibliography{icme2022template}

\begin{thebibliography}{10}

\bibitem{cao2022authentic}
Chen Cao, Tomas Simon, Jin~Kyu Kim, Gabe Schwartz, Michael Zollhoefer,
  Shun-Suke Saito, Stephen Lombardi, Shih-En Wei, Danielle Belko, Shoou-I Yu,
  et~al.,
\newblock ``Authentic volumetric avatars from a phone scan,''
\newblock {\em ACM Transactions on Graphics (TOG)}, vol. 41, no. 4, pp. 1--19,
  2022.

\bibitem{blanz1999morphable}
Volker Blanz and Thomas Vetter,
\newblock ``A morphable model for the synthesis of 3d faces,''
\newblock in {\em Proceedings of the 26th annual conference on Computer
  graphics and interactive techniques}, 1999, pp. 187--194.

\bibitem{zheng2022avatar}
Yufeng Zheng, Victoria~Fern{\'a}ndez Abrevaya, Marcel~C B{\"u}hler, Xu~Chen,
  Michael~J Black, and Otmar Hilliges,
\newblock ``Im avatar: Implicit morphable head avatars from videos,''
\newblock in {\em Proceedings of the IEEE/CVF Conference on Computer Vision and
  Pattern Recognition}, 2022, pp. 13545--13555.

\bibitem{GuyGafni2021DynamicNR}
Guy Gafni, Justus Thies, Michael Zollh{\"o}fer, and Matthias Nie{\ss}ner,
\newblock ``Dynamic neural radiance fields for monocular 4d facial avatar
  reconstruction,''
\newblock {\em computer vision and pattern recognition}, 2021.

\bibitem{AlbertPumarola2020DNeRFNR}
Albert Pumarola, Enric Corona, Gerard Pons-Moll, and Francesc Moreno-Noguer,
\newblock ``D-nerf: Neural radiance fields for dynamic scenes,''
\newblock {\em computer vision and pattern recognition}, 2020.

\bibitem{mildenhall2021nerf}
Ben Mildenhall, Pratul~P Srinivasan, Matthew Tancik, Jonathan~T Barron, Ravi
  Ramamoorthi, and Ren Ng,
\newblock ``Nerf: Representing scenes as neural radiance fields for view
  synthesis,''
\newblock {\em Communications of the ACM}, vol. 65, no. 1, pp. 99--106, 2021.

\bibitem{park2021hypernerf}
Keunhong Park, Utkarsh Sinha, Peter Hedman, Jonathan~T Barron, Sofien Bouaziz,
  Dan~B Goldman, Ricardo Martin-Brualla, and Steven~M Seitz,
\newblock ``Hypernerf: A higher-dimensional representation for topologically
  varying neural radiance fields,''
\newblock {\em arXiv preprint arXiv:2106.13228}, 2021.

\bibitem{park2021nerfies}
Keunhong Park, Utkarsh Sinha, Jonathan~T Barron, Sofien Bouaziz, Dan~B Goldman,
  Steven~M Seitz, and Ricardo Martin-Brualla,
\newblock ``Nerfies: Deformable neural radiance fields,''
\newblock in {\em Proceedings of the IEEE/CVF International Conference on
  Computer Vision}, 2021, pp. 5865--5874.

\bibitem{gecer2019ganfit}
Baris Gecer, Stylianos Ploumpis, Irene Kotsia, and Stefanos Zafeiriou,
\newblock ``Ganfit: Generative adversarial network fitting for high fidelity 3d
  face reconstruction,''
\newblock in {\em Proceedings of the IEEE/CVF conference on computer vision and
  pattern recognition}, 2019, pp. 1155--1164.

\bibitem{feng2021learning}
Yao Feng, Haiwen Feng, Michael~J Black, and Timo Bolkart,
\newblock ``Learning an animatable detailed 3d face model from in-the-wild
  images,''
\newblock {\em ACM Transactions on Graphics (ToG)}, vol. 40, no. 4, pp. 1--13,
  2021.

\bibitem{martin2021nerf}
Ricardo Martin-Brualla, Noha Radwan, Mehdi~SM Sajjadi, Jonathan~T Barron,
  Alexey Dosovitskiy, and Daniel Duckworth,
\newblock ``Nerf in the wild: Neural radiance fields for unconstrained photo
  collections,''
\newblock in {\em Proceedings of the IEEE/CVF Conference on Computer Vision and
  Pattern Recognition}, 2021, pp. 7210--7219.

\bibitem{SongyouPeng2020ConvolutionalON}
Songyou Peng, Michael Niemeyer, Lars Mescheder, Marc Pollefeys, and Andreas
  Geiger,
\newblock ``Convolutional occupancy networks,''
\newblock {\em european conference on computer vision}, 2020.

\bibitem{peng2021neural}
Sida Peng, Yuanqing Zhang, Yinghao Xu, Qianqian Wang, Qing Shuai, Hujun Bao,
  and Xiaowei Zhou,
\newblock ``Neural body: Implicit neural representations with structured latent
  codes for novel view synthesis of dynamic humans,''
\newblock in {\em Proceedings of the IEEE/CVF Conference on Computer Vision and
  Pattern Recognition}, 2021.

\bibitem{BenjaminGraham20173DSS}
Benjamin Graham, Martin Engelcke, and Laurens van~der Maaten,
\newblock ``3d semantic segmentation with submanifold sparse convolutional
  networks,''
\newblock {\em computer vision and pattern recognition}, 2017.

\bibitem{liu2022semantic}
Xian Liu, Yinghao Xu, Qianyi Wu, Hang Zhou, Wayne Wu, and Bolei Zhou,
\newblock ``Semantic-aware implicit neural audio-driven video portrait
  generation,''
\newblock {\em arXiv preprint arXiv:2201.07786}, 2022.

\bibitem{guo2021ad}
Yudong Guo, Keyu Chen, Sen Liang, Yong-Jin Liu, Hujun Bao, and Juyong Zhang,
\newblock ``Ad-nerf: Audio driven neural radiance fields for talking head
  synthesis,''
\newblock in {\em Proceedings of the IEEE/CVF International Conference on
  Computer Vision}, 2021, pp. 5784--5794.

\bibitem{ke2022modnet}
Zhanghan Ke, Jiayu Sun, Kaican Li, Qiong Yan, and Rynson~WH Lau,
\newblock ``Modnet: Real-time trimap-free portrait matting via objective
  decomposition,''
\newblock in {\em Proceedings of the AAAI Conference on Artificial
  Intelligence}, 2022, vol.~36, pp. 1140--1147.

\bibitem{yu2018bisenet}
Changqian Yu, Jingbo Wang, Chao Peng, Changxin Gao, Gang Yu, and Nong Sang,
\newblock ``Bisenet: Bilateral segmentation network for real-time semantic
  segmentation,''
\newblock in {\em Proceedings of the European conference on computer vision
  (ECCV)}, 2018, pp. 325--341.

\bibitem{wang2004image}
Zhou Wang, Alan~C Bovik, Hamid~R Sheikh, and Eero~P Simoncelli,
\newblock ``Image quality assessment: from error visibility to structural
  similarity,''
\newblock {\em IEEE transactions on image processing}, vol. 13, no. 4, pp.
  600--612, 2004.

\bibitem{zhang2018unreasonable}
Richard Zhang, Phillip Isola, Alexei~A Efros, Eli Shechtman, and Oliver Wang,
\newblock ``The unreasonable effectiveness of deep features as a perceptual
  metric,''
\newblock in {\em Proceedings of the IEEE conference on computer vision and
  pattern recognition}, 2018, pp. 586--595.

\end{thebibliography}

\end{document}